\pdfoutput=1

\documentclass[11pt]{article}

\usepackage{acl}

\usepackage{times}
\usepackage{latexsym}

\usepackage[T1]{fontenc}

\usepackage[utf8]{inputenc}
\usepackage{times}
\usepackage{latexsym}
\usepackage{url}
\usepackage{enumitem}
\usepackage{amsmath,amssymb,amsfonts}
\usepackage{algorithmic}
\usepackage{caption}
\usepackage{subcaption}
\usepackage{graphicx}
\usepackage{textcomp}
\usepackage{url}
\usepackage{makecell}
\usepackage{multirow}
\usepackage{titlesec}
\usepackage{booktabs}
\usepackage{array}
\usepackage{microtype}
\usepackage{kotex}
\usepackage{array}
\usepackage{float}
\usepackage{booktabs}
\usepackage{hyperref}       
\usepackage{amsfonts}       
\usepackage{nicefrac}       
\usepackage{xcolor}         
\usepackage{multirow}

\usepackage{kotex}
\usepackage{algorithm}
\usepackage{algorithmic}
\usepackage{todonotes}
\usepackage{multirow}
\usepackage{float}
\usepackage{comment}
\usepackage{todonotes}
\usepackage{caption}
\usepackage{booktabs, makecell, tabularx}

\newcolumntype{L}{>{\raggedright\arraybackslash}X}
\usepackage{siunitx}
\DeclareUnicodeCharacter{0008}{ }
\usepackage{tablefootnote}

\makeatletter
\newcounter{parentsubcaption}
\newenvironment{subsubcaption}
 {\refstepcounter{sub\@captype}%
  \protected@edef\theparentsubcaption{\@nameuse{thesub\@captype}}%
  \setcounter{parentsubcaption}{\value{sub\@captype}}%
  \setcounter{sub\@captype}{0}%
  \@namedef{thesub\@captype}{\theparentsubcaption--\arabic{sub\@captype}}%
  \ignorespaces
}{%
  \setcounter{sub\@captype}{\value{parentsubcaption}}%
  \ignorespacesafterend
}
\makeatother

\newcommand{\thickhline}{%
    \noalign {\ifnum 0=`}\fi \hrule height 1pt
    \futurelet \reserved@a \@xhline
}

\usepackage{comment}
\usepackage{todonotes}
\usepackage{xcolor}

\usepackage{microtype}

\makeatletter
\renewcommand*{\@fnsymbol}[1]{\ensuremath{\ifcase#1\or \dagger\or *\or \ddagger\or
   \mathsection\or \mathparagraph\or \|\or **\or \dagger\dagger
   \or \ddagger\ddagger \else\@ctrerr\fi}}
\makeatother

\title{Language Chameleon: Transformation analysis between languages using Cross-lingual Post-training based on Pre-trained language models}

\author{Suhyune Son$^{1}$\thanks{\hspace*{0.5em}{These authors contributed equally to this work}},  Chanjun Park $^{1,2\dagger}$,  Jungseob Lee$^{1\dagger}$,  Midan Shim$^{3\dagger}$, \\ \textbf{Chanhee Lee} $^{4}$, \textbf{Yoonna Jang} $^{1}$, \textbf{Jaehyung Seo} $^{1}$,  \textbf{Heuiseok Lim} $^{1}$ \thanks{\hspace*{0.5em}{Corresponding author.}}\\
\\
  $^1$ Korea University, $^2$ Upstage, $^3$ Yonsei University, $^4$ Naver Corporation\\
\texttt{ \{ssh5131, bcj1210, omanma1928, seojae777, morleechee, limhseok\}@korea.ac.kr} \\ \texttt{chanjun.park@upstage.ai}, \texttt{midans26@yonsei.ac.kr} \\  \texttt{chanhee.lee@navercorp.com} \\ 
  }

\begin{document}
\maketitle
\begin{abstract}
As pre-trained language models become more resource-demanding, the inequality between resource-rich languages such as English and resource-scarce languages is worsening. This can be attributed to the fact that the amount of available training data in each language follows the power-law distribution, and most of the languages belong to the long tail of the distribution. Some research areas attempt to mitigate this problem. For example, in cross-lingual transfer learning and multilingual training, the goal is to benefit long-tail languages via the knowledge acquired from resource-rich languages. Although being successful, existing work has mainly focused on experimenting on as many languages as possible. As a result, targeted in-depth analysis is mostly absent. In this study, we focus on a single low-resource language and perform extensive evaluation and probing experiments using cross-lingual post-training (XPT). To make the transfer scenario challenging, we choose Korean as the target language, as it is a language isolate and thus shares almost no typology with English. Results show that XPT not only outperforms or performs on par with monolingual models trained with orders of magnitudes more data but also is highly efficient in the transfer process.
\end{abstract}

\section{Introduction}
Recent research on natural language processing (NLP) has been actively conducted in academics and industries based on large-scale data and Transformer \cite{vaswani2017attention}. Since the advent of studies on pre-trained language models (PLMs), such as Foundation model \cite{bommasani2021opportunities}, GPT-3 \cite{brown2020language}, and Prompt tuning \cite{lester2021power, schick2020exploiting, schick2020s}, and theoretical probing \cite{kaplan2020scaling}, the importance of data efficiency has increased in NLP.

The research on Transformer-based PLM has started with BERT \cite{devlin2018bert} and sequentially transitioned into auto-encoder-based: RoBERTa \cite{liu2019roberta}, ELECTRA \cite{clark2020electra}, and XLM \cite{lample2019cross}; decoder-based: GPT1-3 \cite{radford2018improving, radford2019language, brown2020language}; and encoder–decoder-based: MASS \cite{song2019mass}, BART \cite{lewis2019bart}, and T5 \cite{roberts2019exploring}. These advanced PLMs have significantly improved the performance of many downstream tasks in NLP. The above-mentioned studies belong to language representation, wherein the capability of language expression is essential for the model improvement. However, there are three significant limitations in the previous PLM research despite its necessity and importance.

(\romannumeral 1) \textbf{Limitations of Language Diversity}: Recent studies on the language models are based on English. Although there are 303 languages on Wikipedia, most of its documents are written in English. More than half of 154 languages have less than 10,000 documents. More details of the language resource imbalance are mentioned in Appendix~\ref{appendix:language}. Accordingly, researches are mainly conducted in English, and studies on languages other than English have not been actively conducted owing to the limited amount of data. Because the amount of training data is directly related to the performance of language models, the language resource imbalance can be an obstacle to development. Multilingual language models (e.g., mBERT \cite{pires2019multilingual}) can achieve high performance in low-resource languages (LRLs) such as Korean and Swahili by learning multiple languages in one model simultaneously. However, off-the-shelf multilingual language models have many model parameters, and a large amount of data is required for pre-training, resulting in poor effectiveness and efficiency. Furthermore, according to \citet{hu2020xtreme, pires2019multilingual}, recent multilingual models still show unsatisfactory performance on account of the lack of the ability to project representations of all languages into one space.

(\romannumeral 2) \textbf{Limitations of the Models}: Open AI's GPT-3 was trained with 300 billion datasets with 175 billion model parameters. DeepMind also developed Gopher \cite{rae2021scaling}, which contains 280 billion parameters. Furthermore, Google's Switch Transformer \cite{fedus2021switch} is the first model with more than trillion-level model parameters (1.6 trillion parameters). The application of a state-of-the-art model that requires such large amounts of parameters and data to an LRL is difficult, and the reproduction is almost impossible while fulfilling the data amount and parameters required in the original version.

(\romannumeral 3) \textbf{Limitations in Service}: Models with excessive numbers of parameters are almost impossible to implement in real-world services. A service circumstance with sufficient computing power (e.g., GPU) is required to process large-scale data. Because training the model requires large amounts of model parameters and datasets, companies with insufficient resources or computing environments encounter considerable difficulties configuring and improving services using these latest models.

To alleviate the above-mentioned problems, we expand cross-lingual post-training (XPT) \cite{lee2021exploring} to conduct various probing studies. XPT is a method that converts a Transformer-based English pre-trained language model into a model of the target language using post-training. This method has the advantage of easy application to LRLs because it does not require additional pre-training with large amounts of data and model parameters. Therefore, XPT is studied in Korean, a minor language, using the English pre-trained language model with a small amount (4M) of Korean monocorpus data. XPT outperforms the existing Korean language models and multilingual PLM or achieves similar performance.

However, XPT results have not been compared with experimental results based on various NLP subtasks, and no in-depth analysis of training processes and modeling has been performed. In this study, we assess the performance of the XPT method based on KLUE \cite{park2021klue}, which is a benchmark dataset in Korean. We execute post-training on the English RoBERTa model based on the 4M Korean monocorpus. The obtained performance is similar to or better than those of the Korean scratch pre-trained model and multilingual PLM without an additional pre-training step. Moreover, we analyze the XPT-based model performance with respect to parameter initialization, probe the training process, and implement experiments on the point of source language loss through the GLUE benchmark.

\begin{figure*}[t]
\centering
\includegraphics[width=130mm]{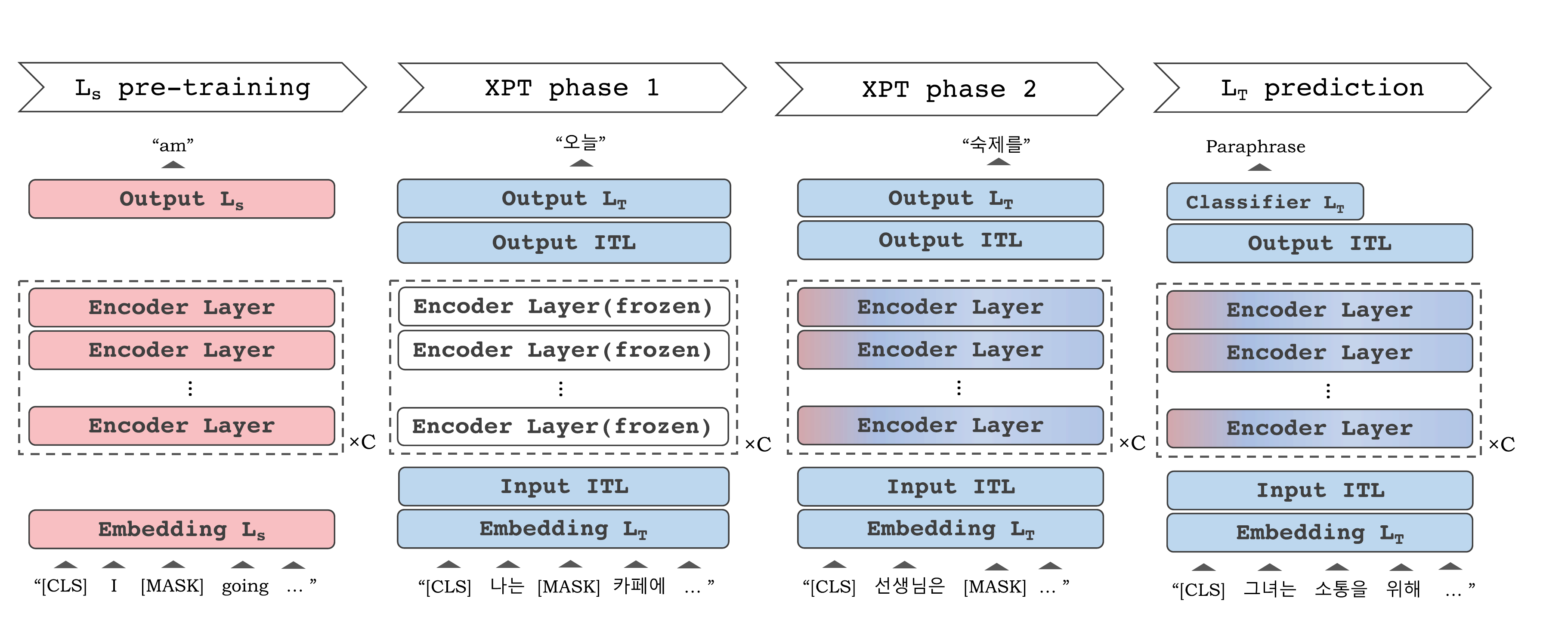}
\caption{Illustration of cross-lingual post-training (XPT) adopted in this study. $C$ is the number of encoder layers. In the input of Phase 1, `나는' means `I' and `카페에' means `at cafe', and in the input of Phase 2, `선생님은' means `teacher'. `그녀는 소통을 위해', the input of $L_{T}$ prediction, means `She $\sim$ for communication'.}
\label{fig:xpt_architecture}
\end{figure*}

\section{Cross-lingual Post-training (XPT)}
\subsection{Overview of XPT}
\label{about_XPT}
A large amount of monolingual corpus is required to pre-train the language model. Existing representative studies of training language models in a low-resource environment are generally divided into the following two approaches: 1) multilingual pre-training and 2) cross lingual transfer learning aiming at a high-to-low-resource language transfer. Among them, XPT belongs to the second category.

XPT performs language transfer using a small amount of the target language corpus in the pre-trained model, as shown in Figure \ref{fig:xpt_architecture}. For this, two phases are passed, and post-training is carried out in English Roberta \cite{liu2019roberta}\footnote{The hyperparameters used in Phase 1 and Phase 2 training are in Appendix \ref{param_xpt}}. Subsequently, the performance is evaluated by fine-tuning the downstream task of the target language $L_T$.

In Phase 1, after freezing all parameters of the encoder layer of the source language model, only the implicit translation layer (ITL) that is additionally added to the input and output of the embedding layer and the embedding layer are trained. Therefore, by freezing the parameters of the encoding layer, only ITLs and embeddings participate in training. Freezing the encoding layer can maintain the parameters well-trained by the existing English model.

In Phase 2, the embedding layers, encoding layers, output layer, and ITLs are all trained. This leverages diluting information that is irrelevant to the target language by performing transfer learning on the encoding layers. Furthermore, this procedure perfectly transforms the English model into the target language model. This post-training process reveals the XPT’s advantage of no need for pre-training with a large-scale target language corpus. This is because XPT trains by appropriately reusing the parameters of the source language model. In addition, XPT utilizes less computational resources than existing models that learn a large-capacity language representation. 
 
Therefore, in this study, we aim to transfer learning from an English model to a Korean model using XPT (the reason for choosing Korean is described in Section \ref{sec:why}). In \citet{lee2021exploring}, the XPT model performance was analyzed in the presence and absence of Phase 1 and the downstream task to the corpus size.

However, various experiments on XPT are still to be conducted, for example, performance evaluation in the presence and absence of ITLs, comparative experiments with various Korean language models, and investigation of the effect of the parameter reuse. To fill this literature gap, in this study, the XPT model is compared with other Korean language models in terms of performance on several Korean downstream tasks. In addition, the effects of a limited corpus environment and reuse of the source language model parameters on the XPT model performance are examined. Further, the exact point of language loss of the source language model is examined, and meaningful analysis results of XPT are derived by probing the performance change according to with and without ITLs.

\subsection{Low Occupied Language (LOL)}
We present a new concept of low occupied language (LOL). In general, low resource langauge (LRL) refers to a language with a small corpus size or a small number of language-speaking populations. We define LOL as a language that has a small corpus size, a low linguistic similarity to English, which is a representative high-resource language, and the characteristics that are not central to language typologies and language isolate. That is, LOL is a language that does not occupy a large portion in terms of linguistics beyond a small amount of data.

We consider Korean to be a representative LOL from the point of view of language typology. Korean is both a minor language and an isolated language, and we intend to conduct various experiments based on these.
\subsection{Korean Language as LOL}
\label{sec:why}
\paragraph{Language Typology}
Linguistic typology is a type of linguistics that classifies language types in consideration of the rules of sentence component, sentence element, and subject ellipsis of a language.

In recent studies of language transfer learning considering the aspect of linguistic typology, it has been confirmed that the more similar the typology between two languages, the better the performance \cite{pires2019multilingual,wang2019cross,ahmad2018difficulties}. In other words, transfer learning between languages with high similarity results in an excellent performance, as shown in several studies \cite{de2020good, de2021adapting, pires2019multilingual}.

However, sufficient studies on transfer learning between languages with low similarity have not been conducted. That is, transfer learning studies on LOL are limited. Therefore, we select Korean as the target language. It belongs to LOLs and has low similarity to English in the aspect of language typology.

\paragraph{Language Isolation} 
In studies on comparative linguistics, language family is defined based on language history and relationships \cite{campbell2017language}. Languages belonging to the same language family share cross-linguistic properties. Among them, Korean is an isolated language with significantly different semantic properties from those of other languages; that is, linguistic commonalities are small \cite{campbell2010language, song2006korean}. Thus, Korean is a language with few linguistic features shared with other languages.

\paragraph{Evaluation Capability}
Although Korean is an LOL, Korean pre-trained language models such as KoBERT\footnote{https://github.com/SKTBrain/KoBERT} and KoELECTRA \cite{park2020koelectra} exist. In addition, there is a benchmark dataset: KLUE \cite{park2021klue}, enabling objective comparison.
Hence, in this study, after training a Korean language model with XPT, a comparison is performed on various existing Korean language models using the benchmark data.

\subsection{Training Process}
\subsubsection{Pre-training}
In this study, pre-training with a large corpus is not required for applying the XPT methodology.

We use pre-trained RoBERTa\footnote{https://github.com/pytorch/fairseq/tree/main/examples/roberta} as the source language model in our approach because it adapts advanced strategies such as dynamic masking and full sentence. 

\subsubsection{Post-training} 
\paragraph{Learning Parameters}
The language model inputs a sequence of tokens $\tau = \left[t_1, t_2, ..., t_n\right]$ and outputs its contextualized representation. Here, $t_i$ denotes the $i$-th token of the input sentence, and $n$ denotes the number of tokens from the tokenizer. In this case, the output of the language model LM can be represented by Equation \ref{final_output}.

\begin{flalign}
    E_{L_S} &= \left[e_1^S;e_2^S;...;e_V^S;\right], e_i^S \in \mathbb{R} ^d \\
    \mathcal{H}_0 &= Embedding(\tau, E ) \\ &=\left[e_{t_1}^S, e_{t_2}^S,..., e_{t_n}^S\right] \\ 
    \mathcal{H}_l &= Encoder_l(\mathcal{H}_{l-1}, \Theta_l ) \label{H_C}\\ &=\left[h_{1}^l, h_{2}^l,..., h_{n}^l\right] \\
    LM(\tau, \Theta) &= \mathcal{H}_C \label{param} \\ &= \left[h_{1}^C, h_{2}^C,..., h_{n}^C\right] \label{final_output}
\end{flalign}

where $E_{L_S}$ is the embedding matrix of $L_S$, $H_l$ is the $l$-th layer's hidden states, and $C$ is the number of the $LM$'s encoder layers. Here, $\Theta$ is $\left[E_{L_S}, \theta_1, \theta_2, ..., \theta_C\right]$. 
 
However, in transfer learning, training all parameters, $\Theta$, is computation-intensive, and there is a risk of diluting the well-trained weights by re-training even the parameters that are helpful for the target language. This means that both meaningful and redundant parameters for the target language are included into $\Theta$ in Equation \ref{param}.
 
In efficient transfer learning, only the parameters of the layers necessary for the target language are trained. To realize this, XPT consists of two phases. In the first phase, layers are separated into useful and redundant for training.
 
Considering this, the pre-trained RoBERTa structure consists of an embedding layer, a hidden layer, and an output layer. In Phase 1, the parameters of the encoding layer are frozen, and the parameters of the word embedding and output layers are initialized to random values and trained. Here, language-independent parameters such as positional embeddings and language modeling head are trained without freeze. Next, in Phase 2, all layers are trained.
 
As a consequence, the embedding layer is trained depending on the vocabulary of the target model, and the encoding layer contextualizes the word vectors projected on the semantic space. The encoding layer is bidirectional. That is, parameters in the encoding layer can be shared between languages and can be reused.

\paragraph{Implicit Translation Layer (ITL)}
Two major limitations are caused by simply freezing the parameters of the encoding layer and by initializing the parameters of the embedding and output layers.

The first problem is word alignment. It is unlikely that a one-to-one correspondence word exists between two languages with low similarity: a source language and a target language. The second problem is linguistic difference. Depending on the grammatical difference between the two languages, the word order and the semantic expression may be different. However, there is a limit in reflecting these differences only by randomly initializing word embeddings and then training to the target. A separate module or layer that can learn these differences is required.
 
Therefore, in this study, we add ITLs, as in \citet{lee2021exploring}, to train differences between the languages. For simplicity and high performance, the ITL consists of a layer with a structure that perfectly matches the structure of the encoding layer \cite{houlsby2019parameter}. $ITL_{in}$ is added between the input layer and the encoding layers, and $ITL_{out}$ is added between the encoding layers and the output layers. The output of the XPT model with ITLs can be represented by Equation \ref{final}.
 
\begin{flalign}
    E_{L_T} &= \left[e_1^T;e_1^T;...;e_1^T;\right], e_i^T \in \mathbb{R} ^d \\
    \mathcal{H}_{L_T} &= Embedding(\tau, E_{L_T} ) \\ 
    \mathcal{H}_0 &= ITL_{in}(\mathcal{H}_{L_T}, \theta_{ITL_{in}})\\
    LM(\tau, \Theta^\prime) &= ITL_{out}(\mathcal{H}_C, \theta_{ITL_{out}}) \label{final}
\end{flalign}

where $E_{L_T}$ is the embedding matrix of the target language model $L_T$, $\tau$ is a sequence of tokens, and $\mathcal{H}_C$ is the same as Equation \ref{H_C}. Because XPT with ITLs is structurally consistent with the previous architecture, it can guarantee the stability of computational and spatial complexities. Accordingly, our method indemnities the efficiency of training.

\subsubsection{Fine-tuning} 
We estimate the model performance by fine-tuning with downstream tasks, similar to existing PLMs such as BERT \cite{devlin2018bert} and RoBERTa. \footnote{The hyperparameters used in fine-tuning are described in the Appendix \ref{param_ft}}
 
We fine-tune the XPT model trained with two phases on the KLUE benchmark and compare it with the existing Korean language and multilingual models. The KLUE benchmark is a Korean NLU task dataset consisting of eight tasks (Topic Classification, Semantic Textual Similarity, Natural Language Inference, Named Entity Recognition, Relation Extraction, Dependency Parsing, Machine Reading Comprehension, and Dialogue State Tracking). The KLUE benchmark provides a convincing criterion for performance measurement in Korean NLU tasks \cite{ban2021survey}. Therefore, we use the KLUE benchmark to evaluate the performance of the XPT model on Korean downstream tasks. 

\section{Experimental Setting}
\subsection{Experimental Design}
We substantiate the validity of the XPT method with the following four experiments. 

First, we perform a fair comparison of our XPT model with other models on KLUE, the Korean benchmark. Several language models are adopted in this experiment: multilingual encoder-based mBERT \cite{pires2019multilingual}, large corpus-based XLM-R \cite{conneau2019unsupervised}, and some Korean language models such as KoELECTRA \cite{park2020koelectra}, KLUE-BERT \cite{park2021klue}, and KLUE-RoBERTa \cite{park2021klue}. KLUE-BERT and KLUE-RoBERTa are proposed in the KLUE benchmark. We perform post-training on two types of corpus sizes: 400K and 4M, to reveal the influence of the corpus size on the performance. These are quite smaller than KoBERT, a representative Korean language model, using 5M and KoELECTRA using 34M.

Second, we compare our model with two KoELECTRAs, trained from scratch on corpus sizes of 400K and 4M, respectively. This experiment is intended to demonstrate the efficiency of the XPT approach that reuses the parameters of the source language. In contrast, the existing methods are newly pre-trained to build a language model with the target language corpus of the same constraint. 

Third, we adapt XPT to a language model, which is randomly initialized RoBERTa. This allows us to probe our approach. We also demonstrate the importance and validity of the design of post-training.

Finally, we ascertain the exact point of language loss. As mentioned in Section \ref{about_XPT}, the XPT approach aims to transfer between languages via two-phase post-training. Accordingly, we perform our experiments to analyze the point of source language loss in each phase. We conduct a quantitative analysis on the GLUE benchmark, which is a well-known English benchmark.

\begin{table*}[ht]

\centering
\scalebox{0.61}{
\renewcommand{\arraystretch}{1.4}

\begin{tabular}{l||c|c|cc|c|cc|cc|cccc|cc}
\hline
     & Corpus      & YNAT                          & \multicolumn{2}{c|}{KLUE-STS}                                                               & KLUE-NLI                               & \multicolumn{2}{c|}{KLUE-NER}                                                      & \multicolumn{2}{c|}{KLUE-RE}                                                                  & \multicolumn{4}{c|}{KLUE-DP}                                                                                                                                                                      & \multicolumn{2}{c}{KLUE-MRC}                                                                                                          \\ \cline{3-16} 
    & Size         & $F1$                            & \multicolumn{1}{c}{$R^P$}                   & $F1$                            & $ACC$                                    & \multicolumn{1}{c}{$F1^E$}         & $F1^C$         & \multicolumn{1}{c}{$F1^{mic}$}         & $AUC$                           & \multicolumn{1}{c}{$UAS^{mac}$}  & \multicolumn{1}{c}{$UAS^{mic}$}  & \multicolumn{1}{c}{$LAS^{mac}$}  & $LAS^{mic}$  & \multicolumn{1}{c}{$EM$}                                                   & $ROUGE$                                                \\ \hline \hline
    
mBERT   & 15G $\uparrow$     & 81.39                         & \multicolumn{1}{c}{84.22}                                  & 77.48                         & 72.57                                  & \multicolumn{1}{c}{75.11}                         & 87.50                         & \multicolumn{1}{c}{ 57.23}       & 58.30                         & \multicolumn{1}{c}{90.59}                        & \multicolumn{1}{c}{92.08}                         & \multicolumn{1}{c}{85.18}                        & 89.61                         & \multicolumn{1}{c}{25.70}                                                & 44.82                                                \\
XLM-R    & 2.5T    & 84.13                         & \multicolumn{1}{c}{88.37}                                  & 81.60                         & 79.03                                  & \multicolumn{1}{c}{80.13}                         & \textbf{91.31}                & \multicolumn{1}{c}{50.10}                               & 52.23                         & \multicolumn{1}{c}{{ \underline {92.90}}}                   & \multicolumn{1}{c}{93.84}                         & \multicolumn{1}{c}{{ \underline{87.18}}}                  & { \underline{91.87}}                   & \multicolumn{1}{c}{28.25} &  54.24 \\
KoELECTRA & 34G   & 84.55                         & \multicolumn{1}{c}{{\underline{90.21}}}                            & 83.67                         & {\underline{83.60}}                            & \multicolumn{1}{c}{66.43}                         & 88.19                        & \multicolumn{1}{c}{57.32}                               &  53.45 & \multicolumn{1}{c}{92.84}                         & \multicolumn{1}{c}{{\underline{94.05}}}                   & \multicolumn{1}{c}{85.51}                        & 91.70                         & \multicolumn{1}{c}{42.44}                                                & 59.72                                                \\
KLUE-BERT & 63G    & \textbf{86.77}                & \multicolumn{1}{c}{89.62}                                  & 82.93                         & 82.33                                  & \multicolumn{1}{c}{\textbf{84.14}}                & 90.58                         & \multicolumn{1}{c}{65.64}                               & {\underline{68.98}}                   & \multicolumn{1}{c}{91.97}                        & \multicolumn{1}{c}{93.34}                        & \multicolumn{1}{c}{84.46}                        & 90.89                        & \multicolumn{1}{c}{{\underline{62.67}}}                                          & {\underline{67.20}}                                          \\
KLUE-RoBERTa & 63G & 85.80 & \multicolumn{1}{c}{ \textbf{92.18}} &  \underline{85.83} &  \textbf{85.70} & \multicolumn{1}{c}{\underline{83.81}} &  90.51 & \multicolumn{1}{c}{ {\underline{66.84}}} &  68.02 & \multicolumn{1}{c}{ 88.99} & \multicolumn{1}{c}{ 92.23} & \multicolumn{1}{c}{ 71.98} &  87.65 & \multicolumn{1}{c}{ \textbf{64.40}}               &  \textbf{79.00}               \\\hline
XPT-400K  & 400K  & 83.23                         & \multicolumn{1}{c}{76.67}                                  & 75.55                         & 64.03                                  & \multicolumn{1}{c}{66.95}                         & 88.53                         & \multicolumn{1}{c}{54.63}                               & 52.08                         & \multicolumn{1}{c}{91.06}                         & \multicolumn{1}{c}{92.24}                         & \multicolumn{1}{c}{85.71}                         & 89.93                         & \multicolumn{1}{c}{26.07}                                                & 48.73                                             \\
XPT-4M & 4M  & {\underline{86.53}}                   & \multicolumn{1}{c}{84.23}                                  & \textbf{90.30}                & 82.03                                  & \multicolumn{1}{c}{70.42}                         & {\underline{91.24}}                   & \multicolumn{1}{c}{\textbf{67.31}}                      & \textbf{72.56}                & \multicolumn{1}{c}{\textbf{93.78}}                & \multicolumn{1}{c}{\textbf{94.61}}                & \multicolumn{1}{c}{\textbf{87.54}}                & \textbf{92.69}  &   \multicolumn{1}{c}{30.39}                                                & 59.18   
 \\ \hline 
\end{tabular}
}
\caption{The comparison results of the XPT model with the Korean/multi-lingual language model for the KLUE dataset and the corpus size. In the case of mBERT trained in the top 104 languages on Wikipedia, the amount of training corpus was not disclosed. However, since the size of the English corpus with the largest amount of resources is 15G, the total amount of corpus used for mBERT training is estimated to be more than 15G. Bold and underline denote the best and the second-best performance, respectively. }
\label{tab:result_koPLM}
\end{table*}

\subsection{Dataset Details}
To train a primitive language model using XPT, a refined Korean dataset is required. First, we extract documents from Wikipedia with Wikiextractor\footnote{\url{https://attardi.github.io/wikiextractor/}} and then separate them into sentences. The final dataset consists of 4.19M sentences. We split it into a 4M training dataset, 100K validation dataset, and 88K test dataset. 

We also utilize two datasets for a fair evaluation of the XPT model. One is the KLUE benchmark for Korean, our target language. Another is the GLUE benchmark to perform fine-tuning in Phases 1 and 2 in order to demonstrate language loss points.
We describe these benchmarks in Appendix \ref{appendix:eval}.

\subsection{Model Details}
We conduct a comparative evaluation of XPT with the models below in terms of performance.

\paragraph{mBERT} mBERT is a 110M size multilingual pre-trained model, which is trained on 15G of the Wikipedia dataset. It uses a Wordpiece tokenizer with a 119,547 vocabulary size. 

\paragraph{XLM-R} XLM-R is a cross-lingual model (XLM), which is trained with commoncrawl corpus\footnote{\url{https://commoncrawl.org/}} containing a hundred types of languages. Unlike previous XLM, it is based on RoBERTa. It has 550M parameters and a vocabulary size of 95K.

\paragraph{KoELECTRA} KoELECTRA is trained on 34GB data containing Korean Wikipedia, Namuwiki\footnote{\url{https://namu.wiki/}}, Korean news, and so on. This model, which is 3112M, uses a Wordpiece tokenizer and 35K vocabulary words. In addition to existing KoELECTRA, we use KoELECTRA-400K and KoELECTRA-4M, which are pre-trained on 400K and 4M, respectively. This allows us to validate our XPT model by comparing training performances for the same corpus.

\paragraph{KLUE-BERT$_{BASE}$} KLUE-BERT uses a trained morpheme-based subword tokenizer. Its size is 110M, and the vocabulary size is 32M. This model is pre-trained on Modu-corpus\footnote{\url{https://corpus.korean.go.kr/}}, Namuwiki, a Korean newspaper, and so on. The total amount of training datasets is 63GB.

\paragraph{KLUE-RoBERTa$_{BASE}$} KLUE-RoBERTa is trained on the same data as KLUE-BERT using RoBERTa's strategies. The size of this model is 110M, and the vocab size is 32M.

\paragraph{XPT} The XPT model proposed herein is based on RoBERTa and is trained with a Korean Wikipedia dataset. We train XPT-400K and XPT-4M so that we compare them with previous Korean language models according to the size of the training corpus. Our XPT uses a SentencePiece tokenizer with a vocabulary size of 50K, and the model size is 130M. To verify the parameter usage of RoBERTa, we also conduct a comparative analysis with our model with randomly initialized RoBERTa, which is post-trained as Phases 1 and 2 of XPT. 

\begin{table*}[ht]

\centering
\scalebox{0.55}{
\renewcommand{\arraystretch}{1.4}
\begin{tabular}{l||c|cc|c|cc|cc|cccc|cc}
\hline
            & YNAT                          & \multicolumn{2}{c|}{KLUE-STS}                                                               & KLUE-NLI                               & \multicolumn{2}{c|}{KLUE-NER}                                                      & \multicolumn{2}{c|}{KLUE-RE}                                                                  & \multicolumn{4}{c|}{KLUE-DP}                                                                                                                                                                      & \multicolumn{2}{c}{KLUE-MRC}                                                                                                          \\ \cline{2-15} 
            & $F1$                            & \multicolumn{1}{c}{$R^P$}                   & $F1$                            & $ACC$                                    & \multicolumn{1}{c}{$F1^E$}         & $F1^C$         & \multicolumn{1}{c}{$F1^{mic}$}         & $AUC$                           & \multicolumn{1}{c}{$UAS^{mac}$}  & \multicolumn{1}{c}{$UAS^{mic}$}  & \multicolumn{1}{c}{$LAS^{mac}$}  & $LAS^{mic}$  & \multicolumn{1}{c}{$EM$}                                                   & $ROUGE $                                               \\ \hline \hline
\begin{tabular}[c]{@{}c@{}}KoELECTRA-400K\end{tabular}  & 4.17  & \multicolumn{1}{c}{7.79}             & 0.00          & 33.33    & \multicolumn{1}{c}{0.00}              & 0.00             & \multicolumn{1}{c}{0.00}     & 3.47  & \multicolumn{1}{c}{83.14}          & \multicolumn{1}{c}{85.82}          & \multicolumn{1}{c}{78.73}          & 81.70          & \multicolumn{1}{c}{5.08}         & 7.87  \\ 
\begin{tabular}[c]{@{}c@{}}KoELECTRA-4M\end{tabular} & 2.40  & \multicolumn{1}{c}{10.21}            & 0.00          & 33.33    & \multicolumn{1}{c}{0.00}              & 0.00                & \multicolumn{1}{c}{0.00}     & 3.15  & \multicolumn{1}{c}{68.52}          & \multicolumn{1}{c}{71.99}          & \multicolumn{1}{c}{55.62}          & 65.75          & \multicolumn{1}{c}{0.00}         & 3.69  \\ \hline
\begin{tabular}[c]{@{}c@{}}XPT-400K\end{tabular}        & 83.23 & \multicolumn{1}{c}{76.67}            & 75.55         & 64.03    & \multicolumn{1}{c}{66.95}             & 88.53                & \multicolumn{1}{c}{54.63}    & 52.08 & \multicolumn{1}{c}{91.06}          & \multicolumn{1}{c}{92.24}          & \multicolumn{1}{c}{85.71}          & 89.93          & \multicolumn{1}{c}{26.07}        & 48.73 \\ 
\begin{tabular}[c]{@{}c@{}}XPT-4M\end{tabular}       & \textbf{86.53} & \multicolumn{1}{c}{\textbf{84.23}}            & \textbf{90.30}          & \textbf{82.03}    & \multicolumn{1}{c}{\textbf{70.42}}             & \textbf{91.24}                & \multicolumn{1}{c}{\textbf{67.31}}    & \textbf{72.56} & \multicolumn{1}{c}{\textbf{93.78}}          & \multicolumn{1}{c}{\textbf{94.61}}          & \multicolumn{1}{c}{\textbf{87.54}}          & \textbf{92.69}          & \multicolumn{1}{c}{\textbf{30.39}}        & \textbf{59.18}  \\ \hline 
\end{tabular}
}
\caption{Experimental results of KoELECTRA-400K, KoELECTRA-4M, XPT-400K, and XPT-4M on KLUE dataset. The number after the model name indicates the size of the Korean corpus used for pre-training. Best performances are highlighted in bold.}
\label{tab:result_KoELEC}
\end{table*}
\begin{table*}[ht]

\centering
\scalebox{0.55}{
\renewcommand{\arraystretch}{1.4}
\begin{tabular}{l||c|cc|c|cc|cc|cccc|cc}
\hline
            & YNAT                          & \multicolumn{2}{c|}{KLUE-STS}                                                               & KLUE-NLI                               & \multicolumn{2}{c|}{KLUE-NER}                                                      & \multicolumn{2}{c|}{KLUE-RE}                                                                  & \multicolumn{4}{c|}{KLUE-DP}                                                                                                                                                                      & \multicolumn{2}{c}{KLUE-MRC}                                                                                                          \\ \cline{2-15} 
            & $F1$                            & \multicolumn{1}{c}{$R^P$}                   & $F1$                            & $ACC$                                    & \multicolumn{1}{c}{$F1^E$}         & $F1^C$         & \multicolumn{1}{c}{$F1^{mic}$}         & $AUC$                           & \multicolumn{1}{c}{$UAS^{mac}$}  & \multicolumn{1}{c}{$UAS^{mic}$}  & \multicolumn{1}{c}{$LAS^{mac}$}  & $LAS^{mic}$  & \multicolumn{1}{c}{$EM$}                                                   & $ROUGE$                                                \\ \hline \hline
\begin{tabular}[c]{@{}l@{}}XPT-4M w/ random \end{tabular} & 85.67     & 77.52    & 75.35                       & 69.63        & 69.28                 & 90.12                   & 47.24           & 44.99   & 92.76              & 93.89             & 86.38           & 91.70      & 24.82       & 47.39  \\ \hline
XPT-4M                                                                              & \textbf{86.53} & \textbf{84.23}                       & \textbf{90.30}                  & \textbf{82.03}   & \textbf{70.42}             & \textbf{91.24}                & \textbf{67.31}        & \textbf{72.56}    & \textbf{93.78}          & \textbf{94.61}          & \textbf{87.54}          & \textbf{92.69}          & \textbf{30.39}          & \textbf{59.18} \\ \hline
\end{tabular}
}
\caption{The comparative results of XPT-4M models with RoBERTa and random initialized RoBERTa. Best performances are highlighted in bold.}
\label{tab:result_random}
\end{table*}

\subsection{Evaluation Details}

We experiment on KLUE and GLUE, which is Korean and English benchmarks, respectively. More detail of our evaluation tasks are presented in Appendix \ref{appendix:eval}.

For all tables, let $\protect R^{P}$ be Pearson’s $\protect \textit{r}$ and $\protect R^{S}$ be Spearman $\protect \textit{r}$. $\protect F1^{E}$, $\protect F1^{C}$, and $\protect F1^{mic}$ denote entity\char`_micro\char`_F1, character\char`_micro\char`_F1, and micro\char`_F1, respectively. Furthermore, $\protect UAS^{mac}$, $\protect UAS^{mic}$, $\protect LAS^{mac}$, and $\protect LAS^{mic}$ denote uas\char`_macro\char`_f1, uas\char`_micro\char`_f1, las\char`_macro\char`_f1, and las\char`_micro\char`_f1, respectively.

\section{Experimental Results}
\subsection{KLUE Benchmark}
We perform comparative experiments on existing Korean pre-trained language models, multilingual models, and XPT model on the KLUE benchmark, which is a representative Korean NLU benchmark dataset. Table \ref{tab:result_koPLM} shows the results.
As a result, XPT-4M trained on 4M of the Korean corpus shows the best or second-best performance on all tasks except for KLUE-NLI and KLUE-MRC. Considering that XPT-4M is trained with a significantly smaller number of data and trainable parameters, we infer that XPT is meaningful just to approximate the performance of the existing models. The results indicate that a language model can be transferred without pre-training it from scratch in a specific language. In other words, PLMs can be changed into the desired language like a chameleon, and good performance can be achieved. 

The XPT-400K model that is trained only with 10\% of the 4M Korean corpus does not outperform the existing models, but it shows a comparable performance to them in all tasks except for KLUE-MRC. The XPT-4M model exhibits higher performance than multilingual models such as mBERT and XLM-R in the KLUE-MRC task but lower performance than Korean pre-trained models such as KLUE-BERT and KLUE-RoBERTa. This result implies that pre-trained models where multiple languages are trained together are likely to affect the MRC task negatively. In addition, this result suggests that the pre-trained model of KLUE may be biased towards the dataset itself.

\begin{table*}[ht]

\centering
\scalebox{0.52}{
\renewcommand{\arraystretch}{1.1}
\begin{tabular}{ccc||cc|c|cc|cc|cc|c|c|c|c|c}
\hline
                          &                         &                                                           & \multicolumn{2}{c|}{CoLA} & SST-2 & \multicolumn{2}{c|}{MRPC}          & \multicolumn{2}{c|}{STS-B}              & \multicolumn{2}{c|}{QQP}           & MNLI-m & MNLI-mm & QNLI  & RTE   & WNLI  \\ \cline{4-17} 
                          &                         &                                                           & $MCC$         & $ACC$         & $ACC$   & \multicolumn{1}{c}{$ACC$}   & $F1$    & \multicolumn{1}{c}{$R^P$} & $R^S$ & \multicolumn{1}{c}{$ACC$}   & $F1$    & $ACC$    & $ACC$     & $ACC$   & $ACC$   & $ACC$   \\ \hline \hline
\multicolumn{3}{c||}{Roberta}                                                                                    & \underline{56.27}       & \underline{82.26}       & \textbf{94.38} & \multicolumn{1}{c}{\underline{86.03}} & \underline{89.91} & \multicolumn{1}{c}{\textbf{90.05}}   & \textbf{89.68}    & \multicolumn{1}{c}{\underline{91.12}} & \underline{88.26} & 87.65  & \underline{87.39}   & \textbf{92.82} & \underline{70.04} & \textbf{56.34} \\ \hline
\multicolumn{1}{c|}{\multirow{3}{*}{Vocab-Ko}} & \multicolumn{1}{c|}{\multirow{3}{*}{Phase1}} & \begin{tabular}[c]{@{}c@{}}XPT\\ w/o ITLs\end{tabular} & 0.00        & 69.13       & 84.06 & \multicolumn{1}{c}{76.96} & 83.74 & \multicolumn{1}{c}{79.79}   & 79.63    & \multicolumn{1}{c}{88.58} & 84.62 & 78.40  & 79.42   & 86.18 & 56.68 & \textbf{56.34} \\ \cline{3-17} 
      \multicolumn{1}{c|}{\multirow{3}{*}{}}                    &    \multicolumn{1}{c|}{\multirow{3}{*}{}}                      & \begin{tabular}[c]{@{}c@{}}XPT\\ w ITLs\end{tabular}   & 0.00        & 69.13       & 85.55 & \multicolumn{1}{c}{84.56} & 89.19 & \multicolumn{1}{c}{83.14}   & 83.01    & \multicolumn{1}{c}{88.98} & 85.04 & 79.02  & 79.99   & 88.03 & 59.21 & \textbf{56.34} \\ \hline
\multicolumn{1}{c|}{\multirow{3}{*}{Vocab-Ko}} & \multicolumn{1}{c|}{\multirow{3}{*}{Phase2}} & \begin{tabular}[c]{@{}c@{}}XPT\\ w/o ITLs\end{tabular} & 0.00        & 69.13       & 85.21 & \multicolumn{1}{c}{80.39} & 86.16 & \multicolumn{1}{c}{80.95}   & 80.77    & \multicolumn{1}{c}{88.56} & 84.61 & 76.96  & 77.39   & 85.83 & 53.79 & \textbf{56.34} \\\cline{3-17} 
     \multicolumn{1}{c|}{\multirow{3}{*}{}}                         &       \multicolumn{1}{c|}{\multirow{3}{*}{}}                   & \begin{tabular}[c]{@{}c@{}}XPT\\ w ITLs\end{tabular}   & -2.07       & 69.03       & 85.55 & \multicolumn{1}{c}{78.92} & 85.67 & \multicolumn{1}{c}{84.18}   & 83.99    & \multicolumn{1}{c}{88.98} & 85.13 & 77.69  & 78.14   & 86.33 & 55.96 & \underline{46.48} \\ \hline
\multicolumn{1}{c|}{\multirow{3}{*}{Vocab-En}} & \multicolumn{1}{c|}{\multirow{3}{*}{Phase1}} & \begin{tabular}[c]{@{}c@{}}XPT\\ w/o ITLs\end{tabular} & \textbf{56.81}       & \textbf{82.45}       & \textbf{94.38} & \multicolumn{1}{c}{84.56} & 88.73 & \multicolumn{1}{c}{\underline{89.58}}   & \underline{89.29}    & \multicolumn{1}{c}{\textbf{91.29}} & \textbf{88.44} & \underline{87.80}  & \textbf{87.52}   & \underline{92.60} & \textbf{73.29} & \textbf{56.34} \\  \cline{3-17} 
      \multicolumn{1}{c|}{\multirow{3}{*}{}}                        &        \multicolumn{1}{c|}{\multirow{3}{*}{}}                  & \begin{tabular}[c]{@{}c@{}}XPT\\ w ITLs\end{tabular}   & 48.89       & 79.58       & \underline{93.46} & \multicolumn{1}{c}{\textbf{87.75}} & \textbf{91.26} & \multicolumn{1}{c}{87.93}   & 87.77    & \multicolumn{1}{c}{91.10} & 88.11 & \textbf{87.94}  & \underline{87.39}   & 92.35 & 65.34 & 42.25 \\ \hline
\multicolumn{1}{c|}{\multirow{3}{*}{Vocab-En}} & \multicolumn{1}{c|}{\multirow{3}{*}{Phase2}} & \begin{tabular}[c]{@{}c@{}}XPT\\ w/o ITLs\end{tabular} & 0.00        & 69.13       & 88.65 & \multicolumn{1}{c}{76.47} & 84.66 & \multicolumn{1}{c}{84.63}   & 84.28    & \multicolumn{1}{c}{89.84} & 86.44 & 82.79  & 83.12   & 88.78 & 53.43 & \textbf{56.34} \\  \cline{3-17} 
        \multicolumn{1}{c|}{\multirow{3}{*}{}}                      &       \multicolumn{1}{c|}{\multirow{3}{*}{}}                   & \begin{tabular}[c]{@{}c@{}}XPT\\ w ITLs\end{tabular}   & 0.00        & 69.13       & 89.33 & \multicolumn{1}{c}{80.64} & 87.07 & \multicolumn{1}{c}{86.60}   & 86.32    & \multicolumn{1}{c}{89.91} & 86.40 & 82.30  & 82.97   & 88.52 & 58.12 & 35.21
                          \\ \hline
\end{tabular}
}
\caption{Experiments to identify the language loss point of the models. The models with or without ITL layers in each phase are evaluated on GLUE benchmark. Comparison by using vocabulary and word embedding between source and target languages is also reported. Bold and underline denote the best and the second-best performance respectively.}
\label{tab:result_glue}
\end{table*}

\subsection{Data Efficiency}
To show the data efficiency of post-training, we compare post-trained XPT with a KoELECTRA that is pre-trained from scratch on the same amount of data, 400K and 4M. As shown in Table \ref{tab:result_KoELEC}, XPT outperforms KoELECTRA with a significant margin in both cases of 400K and 4M of data. This result demonstrates the effectiveness of the XPT methodology where the model takes the parameters from the source language model and carries out a transfer learning. In case of KoELECTRA, the loss does not decrease during pre-training, and it seems that the model is not trained adequately, such as predicting only one label for downstream tasks (A detailed analysis is in Appendix \ref{appendix:error_analysis}). Furthermore, the performance change of KoELECTRA-400K to KoELECTRA-4M is not consistent in all tasks, which indicates that the corpus of 400K and 4M is insufficient for pre-training. Meanwhile, XPT shows a steady improvement in performance with increasing data. Even with a small dataset, XPT shows a superior performance with a large margin. This experimental result shows that pre-training from scratch with insufficient corpus is inefficient. In contrast, the XPT model performance is positively affected even with a small amount of training data, implying the efficiency of XPT. Accordingly, we show that post-training with XPT is better in terms of data efficiency than training a model from scratch in the same corpus limited condition.

\subsection{Advantage of the Pre-trained Parameters}
To demonstrate the validity of the assumption that XPT reuses the parameters of the source language model, we compare RoBERTa, the source language model of XPT, with random initialized RoBERTa. The results are presented in Table \ref{tab:result_random}. When RoBERTa is randomly initialized, there is a performance degradation for all tasks after training on 4M. This means that some contextual representation and semantic information inherent in the source language model encoder are language-independent and helpful in learning the target language. Consequently, these experiments show that XPT accelerates the training of the target language by reusing the pre-trained parameters of the source language model, and the common semantics between languages can be shared even if the alphabets do not overlap at all \cite{nivre2016universal}.

\subsection{Point of the Source Language Loss} 
To analyze the point of the source language loss, we evaluate the performance of the XPT 4M models on the GLUE benchmark in the phase of vocabulary and embedding. Here, language loss means the vanishing language-specific ability. Through this experiment, we attempt to identify the point of the source language loss. We then compare the performance with and without ITLs in order to confirm that the reason for the performance improvement is not related to the ITLs.

As the language of the model and the language of the dataset are different, we divide the experiment into two cases. In the first case, XPT with Korean vocabulary and word embedding, being the vocabulary and embedding of the XPT model, is evaluated. In the second case, XPT with English vocabulary and word embedding, being the vocabulary and embedding of RoBERTa, the source language model, is evaluated. The experimental result is reported in Table \ref{tab:result_glue}. 

The result demonstrates that the Phase 1 model with the English vocabulary and embedding of RoBERTa achieves the best or second-best performance in all tasks. This implies that the loss of the source language, English, occurs in Phase 2.

There is no significant difference in the performance of the models with and without ITLs in each phase. This confirms that the performance change in the downstream tasks is not caused by the increase in the number the parameters due to the addition of the ITLs.
In other words, the performance of XPT is not significantly affected by the ITLs.

However, all types of models show low performance on CoLA. This is due to the characteristics of the data. As reported in \citet{liu2019multi}, CoLA is considered as the most challenging task among the tasks of the GLUE benchmark because the task definition and dataset are relatively unique, as compared to other tasks. As it is comparatively unstable in fine-tuning \cite{han2021robust}, it can be assumed to exhibit relatively low performance.
  
\section{Conclusion}
In this study, we expand a XPT method that improves the performance of the model with a low-resource language by leveraging a PLM with a high-resource language. We transferred to the Korean language model with only a small amount of Korean corpus (400K and 4M) by applying XPT to English PLM RoBERTa. We also compared the XPT models with off-the-shelf Korean PLMs using the KLUE benchmark dataset. As a result, the XPT models outperformed off-the-shelf Korean PLMs pre-trained with a large corpus in most tasks. In addition, we confirmed the performance of XPT method is dependent on the corpus size, use of the source language model parameters, and ITLs. Comparable performance and data efficiency demonstrates the demand for the XPT method in a language environment with insufficient resources. In the future work, we plan to apply the XPT method to the encoder–decoder structure as well to substantiate its generalization. Moreover, we will expand the XPT method to other various low-resource languages for scalability.

\bibliography{anthology,custom}
\bibliographystyle{acl_natbib}

\clearpage
\appendix
\onecolumn

\begin{flushleft}
\textbf{\large Appendix}
\end{flushleft}

\section{Imbalance of Language Resources}
\label{appendix:language}

\begin{figure*}[ht]
\centering
\includegraphics[width=160mm]{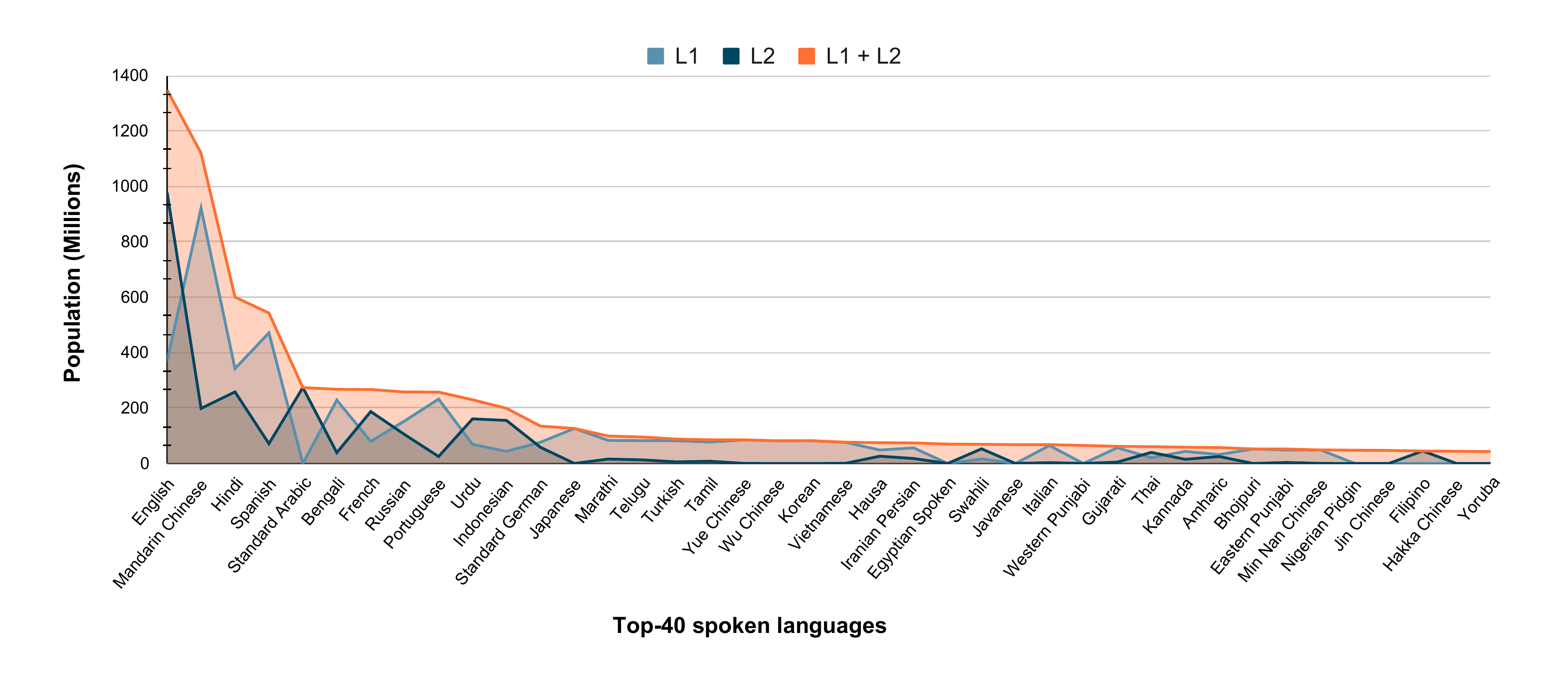}
\caption{Distribution of the population by language for the 40 most spoken languages \footnotemark. L1 denotes the number of native speakers of each language, and L2 denotes the number of second language speakers.}
\label{fig:Population}
\end{figure*}
\footnotetext{\url{https://en.wikipedia.org/wiki/List_of_languages_by_total_number_of_speakers}}
As shown in Figure \ref{fig:Population}, there are 1349 million people who use English as their first and second language. Although various languages exist in the real world, there are many discrepancies in the number of textual resources and the population of languages. We compensate for such an imbalance effectively.

\section{Hyperparameters}
\subsection{Hyperparameters for Post-training}
\label{param_xpt}
\begin{table}[h]

\centering
\scalebox{0.6}{
\renewcommand{\arraystretch}{1.2}

\begin{tabular}{c|c}
\hline
Parameter & Value \\ \hline \hline
Total Steps           & 100K                 \\ \hline
Batch Size            & 512                  \\\hline
Learning Rate         & 3e-4                 \\\hline
Optimizer             & Adam                \\\hline
\end{tabular}}
\caption{XPT training hyperparameters}
\label{tab:xpt_hyper}
\end{table} 

The hyperparameters used for transfer learning are shown in Table \ref{tab:xpt_hyper}.

\subsection{Hyperparameters for Fine-tuning }
\label{param_ft}

\begin{table*}[ht]

\centering
\scalebox{0.6}{
\renewcommand{\arraystretch}{1.2}

\begin{tabular}{c||ccccccc}
\hline
           Hyperparameters     & YNAT & KLUE-STS & KLUE-NLI & KLUE-NER & KLUE-RE & KLUE-DP & KLUE-MRC \\ \hline \hline
Training Epochs & 4    & 4                                & 5        & 3        & 3       & 10      & 5        \\ \hline
Batch Size      & 32   & 32                               & 32       & 32       & 32      & 32      & 16       \\ \hline
Learning Rate   & 5e-5 & 5e-5                             & 5e-5     & 5e-5     & 5e-5    & 5e-5    & 3e-5     \\ \hline
Warmup Ratio    & 0.1  & 0.1                              & 0.1      & None     & 0.1     & 0.1     & None   \\\hline 
\end{tabular}}
\caption{KLUE finetuning hyperparameters}
\label{tab:klue_hyper}
\end{table*}


The hyperparameters used for fine-tuning with the KLUE benchmark are shown in Table \ref{tab:klue_hyper}.

For all GLUE benchmark \cite{wang2018glue} experiments, epoch, batch size, and learning rate are set to 5, 64, and 5e-5, respectively. The dropout ratio is set to 0.3 and 0.05 for the MNLI and CoLA tasks, respectively, and to 0.1 for all other tasks, as in \citet{liu2019multi}.


\begin{table*}[ht]
\centering
\scalebox{0.6}{
\renewcommand{\arraystretch}{1}
\begin{tabular}{l|llllllll}
\hline
\textbf{Dataset} & \textbf{Name} & \textbf{Type}                                                                & \textbf{Format}                                                                                 & \textbf{Eval. Metric}                               & \textbf{\# Class}                                      & \textbf{|Train|, |Dev|, |Test|} & \textbf{Source}                                                                                 & \textbf{Style}                                               \\ \hline \hline
              & YNAT       & \begin{tabular}[c]{@{}l@{}}Topic\\ Classification\end{tabular}               & \begin{tabular}[c]{@{}l@{}}Single Sentence\\ Classification\end{tabular}                        & Macro F1                                                                                       & 7                                                      & 45k, 9k, 9k                       & \begin{tabular}[c]{@{}l@{}}News\\ (Headline)\end{tabular}                                       & Formal                                                       \\ \cline{2-9} 
              & KLUE-STS      & \begin{tabular}[c]{@{}l@{}}Semantic \\ Textual \\ Similarity\end{tabular}    & \begin{tabular}[c]{@{}l@{}}Sentence Pair\\ Classification\end{tabular}                              & \begin{tabular}[c]{@{}l@{}}Person's r, \\ F1\end{tabular}                                      & \begin{tabular}[c]{@{}l@{}}{[}0,5{]} \\ 2\end{tabular} & 11k, 0.5k, 1k                         & \begin{tabular}[c]{@{}l@{}}New,\\ Review,\\ Query\end{tabular}                                  & \begin{tabular}[c]{@{}l@{}}Colloquial,\\ Formal\end{tabular} \\ \cline{2-9} 
              & KLUE-NLI      & \begin{tabular}[c]{@{}l@{}}Natural \\ Language\\ Inference\end{tabular}      & \begin{tabular}[c]{@{}l@{}}Sentence Pair\\ Classification\end{tabular}                          & Accuracy                                                                                       & 3                                                      & 25k, 3k, 3k                           & \begin{tabular}[c]{@{}l@{}}New,\\ Wikipedia,\\ Review\end{tabular}                              & \begin{tabular}[c]{@{}l@{}}Colloquial,\\ Formal\end{tabular} \\ \cline{2-9} 
KLUE          & KLUE-NER      & \begin{tabular}[c]{@{}l@{}}Named\\ Entity\\ Recognition\end{tabular}         & \begin{tabular}[c]{@{}l@{}}Sequence\\ Tagging\end{tabular}                                      & \begin{tabular}[c]{@{}l@{}}Entity-level Macro F1,\\ Character-level Macro F1\end{tabular}      & \begin{tabular}[c]{@{}l@{}}6, \\ 12\end{tabular}       & 21k, 5k, 5k                          & \begin{tabular}[c]{@{}l@{}}News,\\ Review\end{tabular}                                          & \begin{tabular}[c]{@{}l@{}}Colloquial,\\ Formal\end{tabular} \\ \cline{2-9} 
              & KLUE-RE       & \begin{tabular}[c]{@{}l@{}}Relation\\ Extraction\end{tabular}                & \begin{tabular}[c]{@{}l@{}}Single Sentence \\ Classification\\ (with Entity Spans)\end{tabular} & \begin{tabular}[c]{@{}l@{}}Micro F1(without no\_relation), \\ AUPRC\end{tabular}               & 30                                                     & 32k, 8k, 8k                        & \begin{tabular}[c]{@{}l@{}}Wikipedia, \\ News\end{tabular}                                      & Formal                                                       \\ \cline{2-9} 
              & KLUE-DP       & \begin{tabular}[c]{@{}l@{}}Dependeny\\ Parsing\end{tabular}                  & \begin{tabular}[c]{@{}l@{}}Sequence\\ Tagging\\ (with POS tag)\end{tabular}                     & \begin{tabular}[c]{@{}l@{}}Unlabeled Attachment Score,\\ Labeled Attachment Score\end{tabular} & \begin{tabular}[c]{@{}l@{}}\# Words,\\ 38\end{tabular} & 10k, 2k, 2.5k                         & \begin{tabular}[c]{@{}l@{}}News,\\ Review\end{tabular}                                          & \begin{tabular}[c]{@{}l@{}}Colloquial,\\ Formal\end{tabular} \\ \cline{2-9} 
              & KLUE-MRC      & \begin{tabular}[c]{@{}l@{}}Machine\\ Reading\\ Comprehension\end{tabular}    & Span Prediction                                                                                 & \begin{tabular}[c]{@{}l@{}}Exact Match,\\ ROUGE-W (LCCS-based F1)\end{tabular}                 & 2                                                      & 12k, 8k, 9k                      & \begin{tabular}[c]{@{}l@{}}Wikipedia,\\ News\end{tabular}                                       & Formal                                                       \\ \hline \hline
              & CoLA          & \begin{tabular}[c]{@{}l@{}}Linguistic\\ Acceptability\end{tabular}           & \begin{tabular}[c]{@{}l@{}}Sentence\\ Classification\end{tabular}                               & Matthews Correlation                                                                           & 2                                                      & 8.5k, 1k, 1.2k                    & \begin{tabular}[c]{@{}l@{}}Linguistics \\ Literature\end{tabular}                               & Formal                                                       \\ \cline{2-9} 
              & SST-2         & \begin{tabular}[c]{@{}l@{}}Sentimental\\ Analysis\end{tabular}               & \begin{tabular}[c]{@{}l@{}}Sentences\\ Classification\end{tabular}                              & Accuracy                                                                                       & {[}0,5{]}                                              & 67k, 0.9k, 1.8k                  & Movie Review                                                                                    & Colloquial                                                   \\ \cline{2-9} 
              & MRPC          & \begin{tabular}[c]{@{}l@{}}Semantic \\ Textual \\ Similarity\end{tabular}    & \begin{tabular}[c]{@{}l@{}}Sentence Pair\\ Classification\end{tabular}                              & \begin{tabular}[c]{@{}l@{}}Accuracy,\\ F1\end{tabular}                                         & 2                                                      & 3.7k, 0.4k, 1.7k                      & News                                                                                            & Formal                                                       \\ \cline{2-9} 
              & STS-B         & \begin{tabular}[c]{@{}l@{}}Semantic \\ Textual \\ Similarity\end{tabular}    & \begin{tabular}[c]{@{}l@{}}Sentence Pair\\ Regression\end{tabular}                              & \begin{tabular}[c]{@{}l@{}}Pearson,\\ Spearman\end{tabular}                                    & {[}0,5{]}                                              & 5.8k, 1.5k, 1.4k                   & News                                                                                            & Formal                                                       \\ \cline{2-9} 
GLUE          & QQP           & \begin{tabular}[c]{@{}l@{}}Question Pair\\ Similarity\end{tabular}           & \begin{tabular}[c]{@{}l@{}}Sentence Pair\\ Classification\end{tabular}                          & \begin{tabular}[c]{@{}l@{}}Accuracy,\\ F1\end{tabular}                                         & 2                                                      & 364k, 40k, 391k                     & \begin{tabular}[c]{@{}l@{}}Social QA \\ questions\end{tabular}                                  & Colloquial                                                   \\ \cline{2-9} 
              & MNLI-m        & \begin{tabular}[c]{@{}l@{}}Natural \\ Language\\ Inference\end{tabular}      & \begin{tabular}[c]{@{}l@{}}Sentence Pair\\ Classification\end{tabular}                          & Accuracy                                                                                       & 3                                                      & 393k, 20k, 20k                        & \begin{tabular}[c]{@{}l@{}}Fiction,\\ Letter,\\ Government,\\ Travel,\\ Slate, etc\end{tabular} & \begin{tabular}[c]{@{}l@{}}Colloquial,\\ Formal\end{tabular} \\ \cline{2-9} 
              & QNLI          & \begin{tabular}[c]{@{}l@{}}Natural \\ Language\\ Inference\end{tabular}      & \begin{tabular}[c]{@{}l@{}}Sentence Pair\\ Classification\end{tabular}                          & Accuracy                                                                                       & 2                                                      & 105k, 5.5k, 5.5k                     & Wikipedia                                                                                       & Formal                                                       \\ \cline{2-9} 
              & RTE           & \begin{tabular}[c]{@{}l@{}}Recognizing \\ Textual \\ Entailment\end{tabular} & \begin{tabular}[c]{@{}l@{}}Sentences\\ Classification\end{tabular}                             & Accuracy                                                                                       & 2                                                      & 2.5k, 0.3k, 3k                  & \begin{tabular}[c]{@{}l@{}}Wikipedia,\\ News\end{tabular}                                       & Formal                                                       \\ \cline{2-9} 
              & WNLI          & \begin{tabular}[c]{@{}l@{}}Natural \\ Language\\ Inference\end{tabular}      & \begin{tabular}[c]{@{}l@{}}Sentence Pair\\ Classification\end{tabular}                          & Accuracy                                                                                       & 2                                                      & 634, 71, 146                          & Fiction books                                                                                   & Colloquial                                                   \\ \hline
\end{tabular}}

\label{tab:eval}
\caption{\label{tab:eval} The details of the KLUE and GLUE benchmark. }

\end{table*}


\section{Descriptions of KLUE and GLUE Benchmark} 
\label{appendix:eval}
The KLUE benchmark \cite{park2021klue} is used to compare the Korean understanding performance of models, and the GLUE benchmark is utilized to verify the points of source language loss. The details of the two benchmarks are provided in Table \ref{tab:eval}.

Depending on the task of the benchmark, each of them is categorized into four types: sentence classification, sentence regression, range prediction, and sequence tagging. Likewise, the KLUE benchmark consists of sentence classification (KLUE-YNAT, KLUE-NLI, and KLUE-RE), sentence regression (KLUE-STS), sequence tagging (KLUE-NER and KLUE-DP), and span prediction (KLUE-MRC).

\paragraph{KLUE-YNAT} KLUE-YNAT is a multi-label classification task that predicts the topic of the text as one of the seven topics.

\paragraph{KLUE-STS} KLUE-STS is a task that predicts the similarity score (0\texttildelow5) of given sentence pair.

\paragraph{KLUE-NLI} KLUE-NLI is a task that predicts whether the premise entails the hypothesis (entailment), contradicts the hypothesis (contradiction), or neither (neutral) when the premise and hypothesis are given.  

\paragraph{KLUE-NER} KLUE-NER is a token-level tagging task that classifies each token into six named-entity categories.

\paragraph{KLUE-RE} KLUE-RE task classifies the relation between the head entity and tail entity into one of 30 relations in a given sentence. 

\paragraph{KLUE-DP} KLUE-DP is a sequence tagging task that identifies the relation between tokens to analyze the sentence structure. 

\paragraph{KLUE-MRC} KLUE-MRC is a task of predicting the answer span given the context and the question.

\paragraph{KLUE-DST} KLUE-DST is a task that predicts dialogue states in a given dialogue context. For this, It requires a slot generation for prediction. Since it is different from the model structure of this paper based on the encoder structure model, we exclude it from the evaluation. 

The GLUE benchmark consists of STS-B, which is a sentence regression, and the rest of the sentence classification tasks. Like KLUE-STS, STS-B is a task of evaluation of the semantic similarity, as 1\texttildelow5. MRPC and QQP are to predict whether a given sentence pair is semantically identical, binarily. CoLA is a task of determining whether a given sentence is grammatically correct. In addition, RTE is a task of binary classification of the textual entailment. In MNLI tasks, according to a given premise, sentences are classified into three types, like in KLUE-NLI. QNLI uses data which is transformed from the SQuAD dataset to NLI, and it is a task to compare a question and a sentence in a paragraph and to determine whether these two are entailed.

\section{Related Work}
Training low-resource language model is commonly divided into three approaches. The first approach is training large token embeddings with other multiple languages. The second approach is to training token embedding layers with other monolingual language. The last approach is adding a new layer, for training linguistic differences, such as adapters \cite{houlsby2019parameter}.

\paragraph{Training Multilingual Lexical Embeddings}
To facilitate the improvement of the low-resource language model, several studies conducted joint-learning of multilingual embedding layer \cite{devlin2018bert, lample2019cross, conneau2019unsupervised, ri2021mluke,huang2019unicoder}. Advanced performance was observed in a model which contains large embeddings when training deep representation of multilingual language using their large corpus. However, since it uses various languages, performance decline was observed because of an imbalance in the amount of dataset \cite{artetxe2019cross}.

\paragraph{Training Monolingual Lexical Embeddings}
Transfer learning between two languages with high similarity enables language conversion by simply re-training embedding layers. \citet{de2021adapting} transferred monolingual BERT to the Gronings, which is one of the low-resource languages, with re-training only the lexical layer. \citet{de2020good} aligned English and target language by replacing lexical embedding of GPT-2, the generative model. Nevertheless, these are only able to conduct only linguistically similar scenarios where the order and the alignment of word are guaranteed.

\paragraph{Training Additional Layers and Lexical Embeddings}
Recently, there are studies aiming to adapt the pre-trained model to other languages by adding a new layer for language transfer.
\citet{artetxe2019cross} trained target language representation while maintaining pre-trained parameters of source language model only by training language adapter in the model. To alleviate problems in the multilingual model, \citet{pfeiffer2020mad} added language-specific adapters, which leverage effective language transfer to the target, and invertible adapters. This is also aimed at language transfer for unseen languages. 

There are problems that it is dependent only on the additional layer to train the difference between source and target language in existing studies. Since the trained model tends to retain previous parameters, it is natural to the drawback that the model has a deficiency, impossibleness to adapt the linguistic features of the target language.

\section{Error Analysis}
\label{appendix:error_analysis}
\begin{figure*}
\centering
    \begin{subsubcaption}
        \begin{subfigure}{7cm}
        \includegraphics[width=\linewidth]{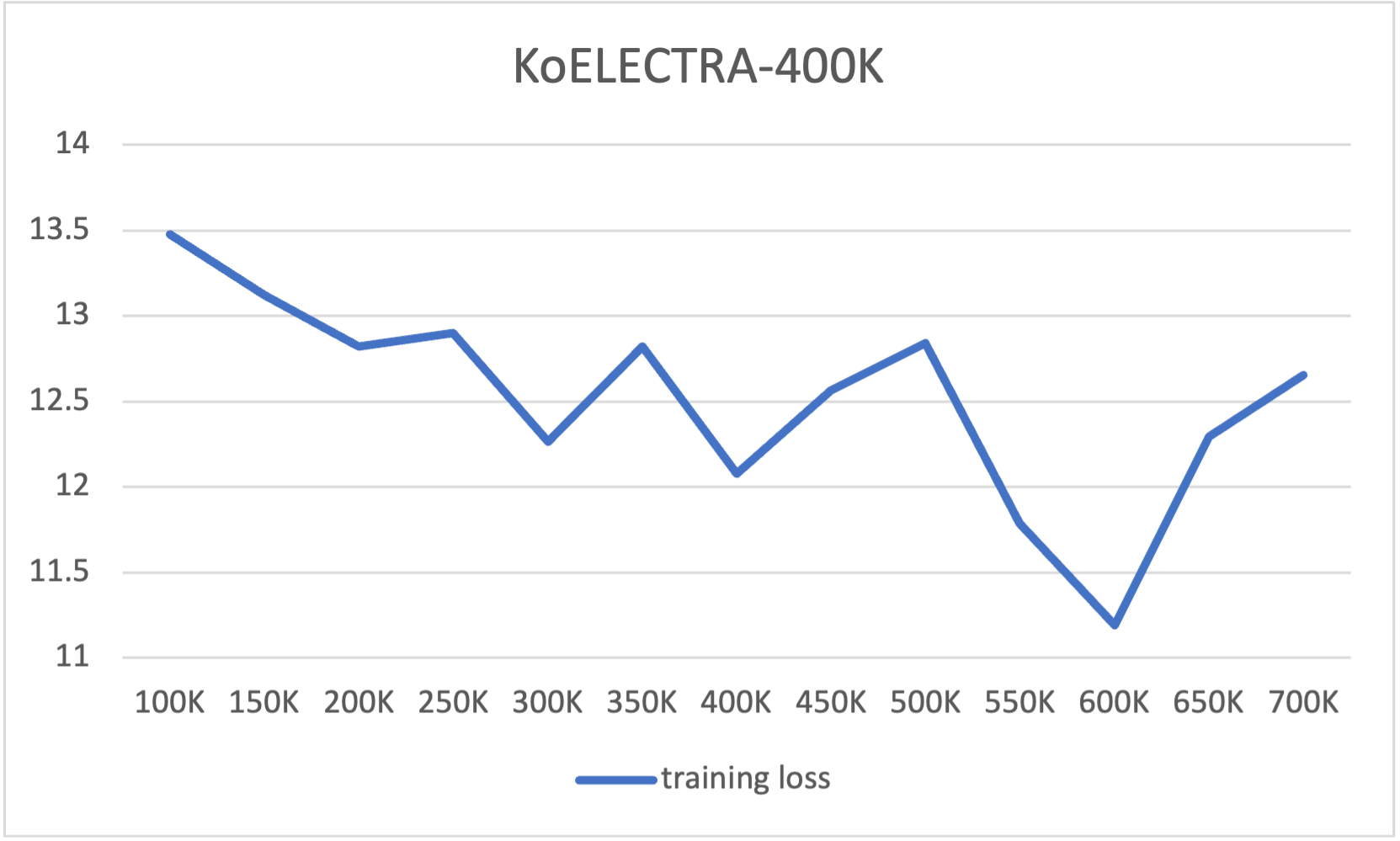}
        \caption{ Loss graph of KoELECTRA-400K} \label{fig:loss_graph-a}
        \end{subfigure}\qquad
        \begin{subfigure}{7cm}
        \includegraphics[width=\linewidth]{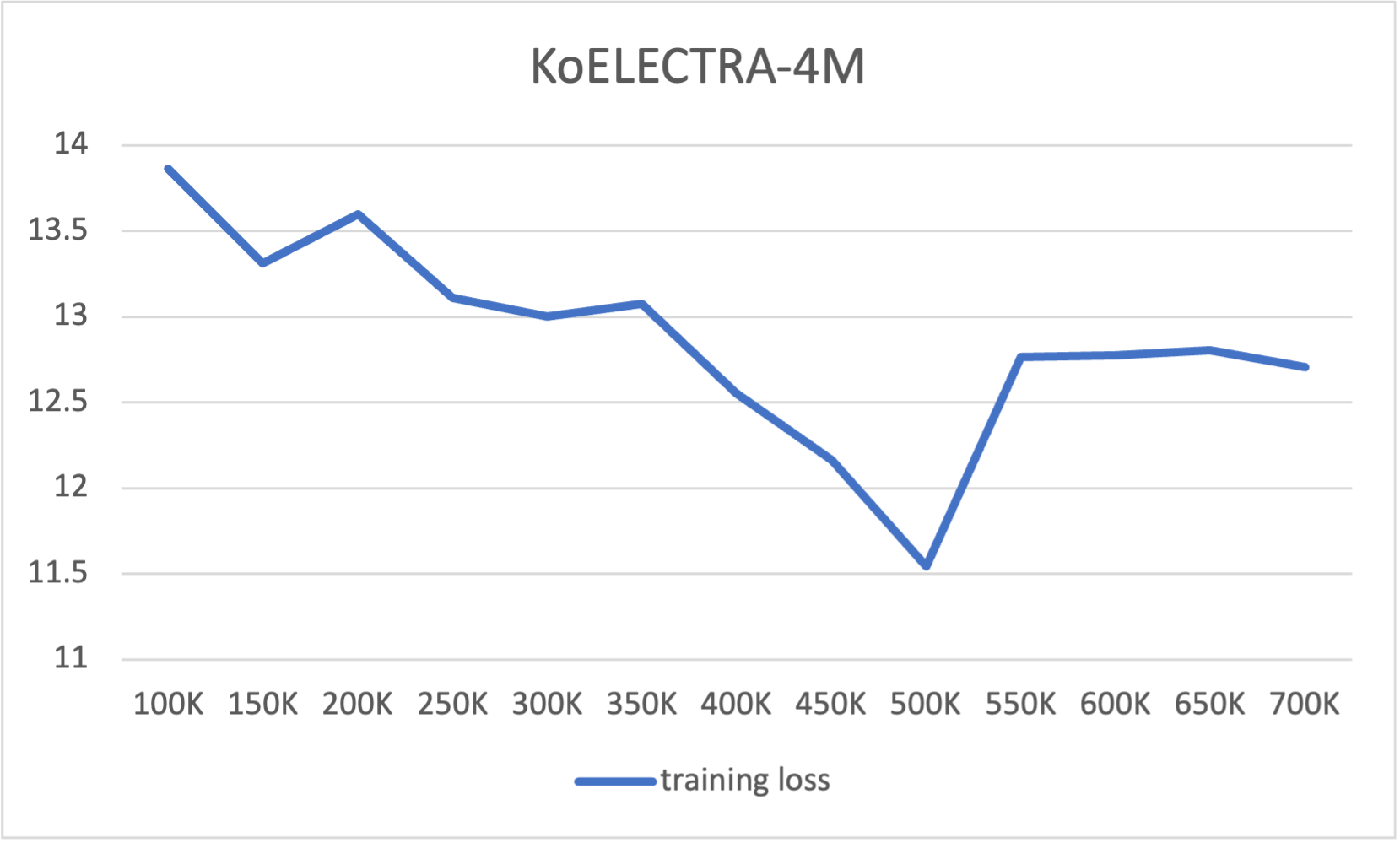}
        \caption{ Loss graph of KoELECTRA-4M}  \label{fig:loss_graph-b}
        \end{subfigure}
        \end{subsubcaption}
    \begin{subsubcaption}
        \begin{subfigure}{7cm}
        \includegraphics[width=\linewidth]{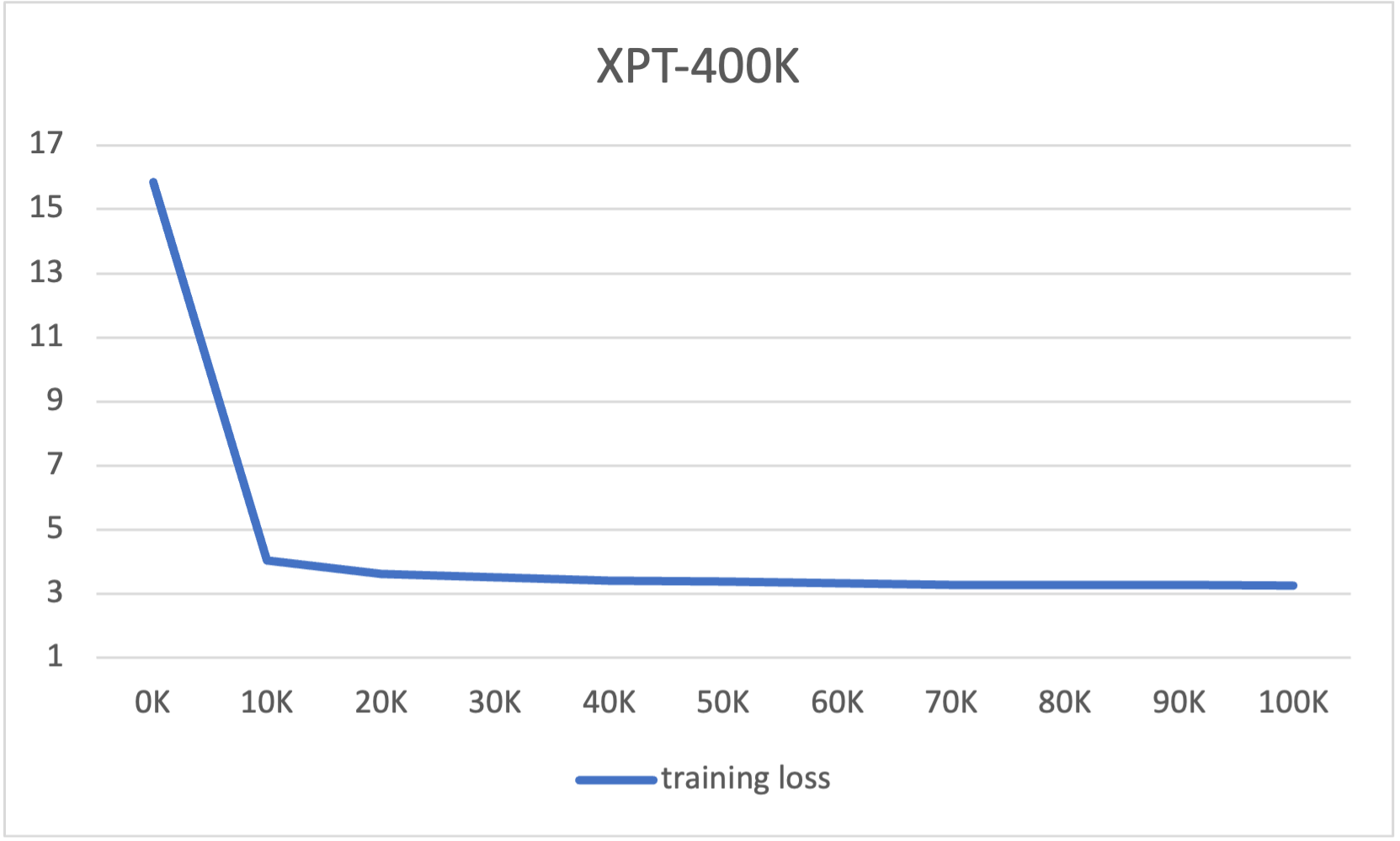}
        \caption{ Loss graph of XPT-400K}  \label{fig:loss_graph-c}
        \end{subfigure}\qquad
        \begin{subfigure}{7cm}
        \includegraphics[width=\linewidth]{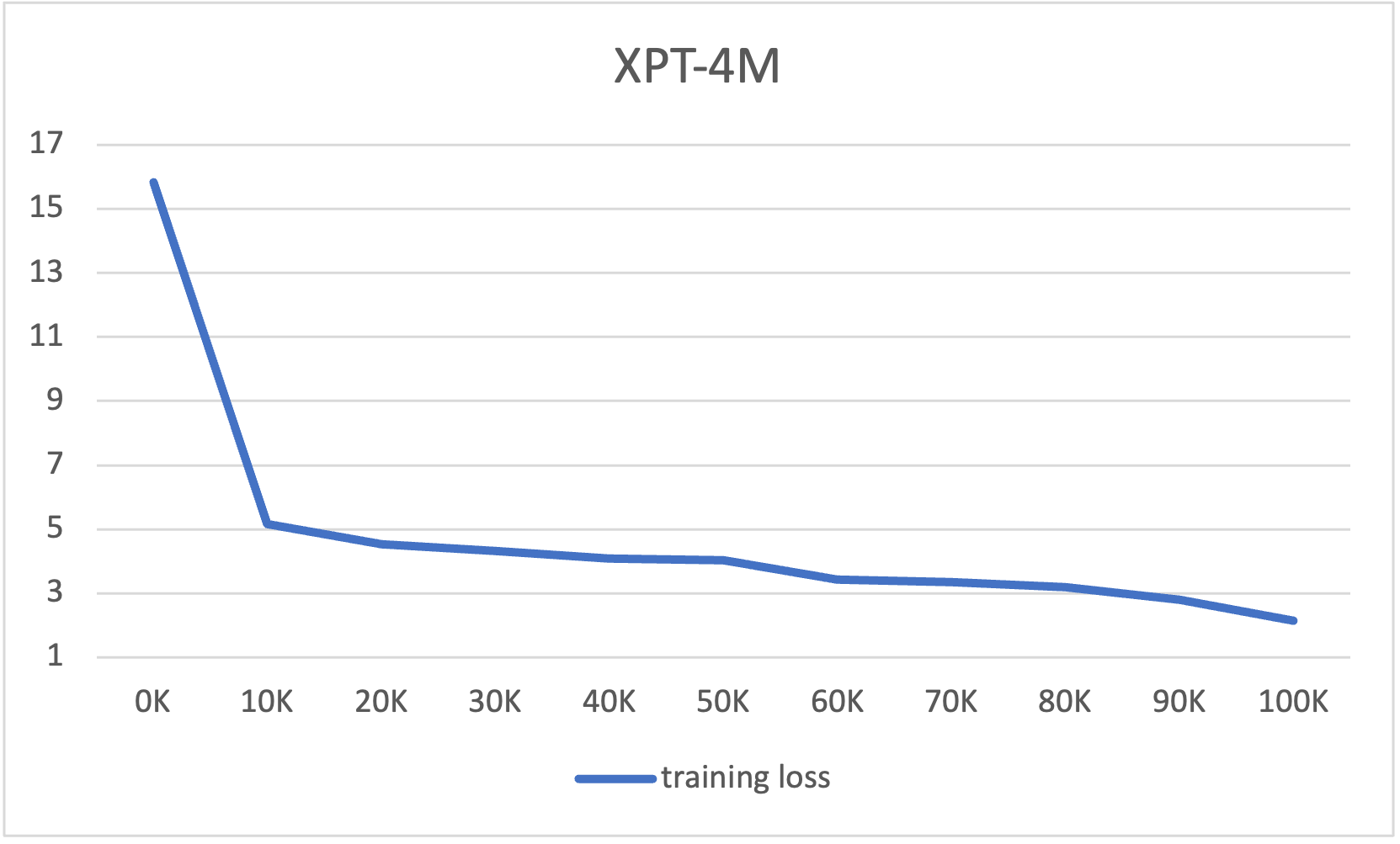}
        \caption{The Loss graph of XPT-4M}  \label{fig:loss_graph-d}
        \end{subfigure}
        \end{subsubcaption}
    \caption{ Loss graph of KoELECTRA-400K, KoELECTRA-4M, XPT-400K, and XPT-4M. The blue line indicates training loss, X-axis represents each training step, and Y-axis represents the loss.}
    \label{fig:loss_graph}
    \end{figure*}

\paragraph{Training Loss of Models}
\label{appendix:model_loss}
Figure \ref{fig:loss_graph} represents the training loss graphs of KoELECTRA-400K, KoELECTRA-4M, XPT-400K, and XPT-4M. The training losses of XPT-400K and XPT-4M gradually decrease as the number of training steps is increased. In general, the training loss decreases as the model training progresses. In contrast, in KoELECTRA-400K and KoELECTRA-4M, the training loss does not decrease uniformly. This indicates that the corpus of 400K and 4M is insufficient for pre-training the language model. That is, KoELECTRA-400K and KoELECTRA-4M, which are not trained fully owing to the small corpus size, show lower performances in downstream tasks. However, because of the characteristic that reuses the parameters of the source language model in post-training, XPT models show a 
desirable loss graph even with the same amount of corpus.

\paragraph{Qualitative Analysis of KLUE-STS} 
\label{appendix:klue-sts}
Qualitative analysis result for KLUE-STS task is presented in Table \ref{appendixtab:sts}. KoELECTRA-400K and KoELECTRA-4M, which are trained from scratch with two sizes of Korean corpus (400K and 4M), predict only one label, ``0.'' In contrast, the XPT models show improved performance with increased corpus size.

\paragraph{Qualitative Analysis of KLUE-NER} 
\label{appendix:klue-ner}
KoELECTRA-400K and KoELECTRA-4M predict only one label, ``O'', as shown in Table \ref{appendixtab:ner}. As we evaluate the performance excluding labels ``O'', as in previous NER studies, the f1 score is 0. The XPT models confuse ``PS (Person)'', ``OG (Organization)'' and ``LC (Location)''. This is because the three labels, ``PS'', ``OG'' and ``LC'' , belong to the same category which is ``entity name expression type'' among the NER types defined in \citet{dahan2015first}. Further, XPT confuses ``QT (Quantity)'', ``DT (Date)'', and ``TI (Time)'' which are related to numbers. That is, the XPT models tend to confuse labels belonging to the same type, similar to previous language models \cite{ouchi2020instance}. 
 
\begin{table}[ht]

\centering
\scalebox{0.65}{
\renewcommand{\arraystretch}{1.2}

\begin{tabular}{c||c|c}
\hline
                   Gold  label     & \textbf{0}                     & \textbf{1}                     \\ \hline
 The number of samples & 299                   & 220                   \\ \hline \hline
KoELECTRA-400K        & \{"0": 299\}          & \{"0": 220\}          \\ \hline
KoELECTRA-4M          & \{"0": 299\}          & \{"0": 220\}          \\ \hline
XPT-400K              & \{"0": 224, "1": 75\} & \{"1": 197, "0": 23\} \\ \hline
XPT-4M                & \{"0": 233, "1": 66\} & \{"1": 207, "0": 13\} \\ \hline
\end{tabular}
}
\caption{Model prediction results for each label in the KLUE-STS task of each model. The number below the gold label means the number of samples tagged with each label in the dev dataset.}
\label{appendixtab:sts}
\end{table}

\paragraph{Qualitative analysis of KLUE-RE} 
\label{appendix:klue-re}
As shown in Table \ref{appendixtab:re}, KoELECTRA-400K and KoELECTRA-4M predict only one relation, that is ``no\_relation''. We evaluate the performance excluding “no\_relation,” as in previous RE studies. The XPT models confuse ``per:title'' with ``org:top\_members/employees.'' The reason for this error is that those two relation definitions are similar, as in Table \ref{appendixtab:relation}. These two relations vary depending on the class (per or org) of sub and obj. In ``per:employee\_of'', ``per:origin'' has the most errors. As in the description of the two relations, the two labels have similar meanings. ``org:product'' is to be confused with ``org:members''. In other words, as both relations define something belonging to an organization, they can be considered to be similar. Consequently, the XPT models confuse similar relations and show reasonable error statics, unlike KoELECTRA.

\begin{table*}[ht]

\centering
\scalebox{0.56}{
\renewcommand{\arraystretch}{1.2}

\begin{tabular}{c||c|c|c|c|c|c}
\hline
\multicolumn{1}{c||}{Gold  label}                   & \textbf{PS (Person)}                                                                                                    & \textbf{QT (Quantity)}                                                                                                               & \textbf{OG (Organization)}                                                                                                            & \textbf{DT (Date)}                                                                                                                   & \textbf{LC (Location)}                                                                                                              & \textbf{TI (Time)}                                                                                \\ \hline
 The number of entities& 13694                                                                                                                     & 10198                                                                                                                      & 9309                                                                                                                        & 8487                                                                                                                       & 6502                                                                                                                      & 2225                                                                                    \\ \hline  \hline
KoELECTRA-400K                            & \{"O": 13694\}                                                                                                            & \{"O": 10198\}                                                                                                             & \{"O": 9309\}                                                                                                               & \{"O": 8487\}                                                                                                              & \{"O": 6502\}                                                                                                             & \{"O": 2225\}                                                                           \\ \hline
KoELECTRA-4M                              & \{"O": 13694\}                                                                                                            & \{"O": 10198\}                                                                                                             & \{"O": 9309\}                                                                                                               & \{"O": 8487\}                                                                                                              & \{"O": 6502\}                                                                                                             & \{"O": 2225\}                                                                           \\ \hline
XPT-400K                                  & \begin{tabular}[c]{@{}c@{}}\{"PS": 12847, \\ "O": 762, "OG": 51, \\ "LC": 15, "QT": 14,\\ "DT":5, "TI": 0 \}\end{tabular} & \begin{tabular}[c]{@{}c@{}}\{"QT": 9742, "O": 314,\\  "DT": 69, "TI": 46, \\ "LC": 9, "OG": 10, \\ "PS": 8 \}\end{tabular} & \begin{tabular}[c]{@{}c@{}}\{"OG": 7995, "O": 825, \\ "LC": 361, "PS": 114, \\ "DT": 10, "QT": 4, \\ "TI":0 \}\end{tabular} & \begin{tabular}[c]{@{}c@{}}\{"DT": 8003, "O": 400, \\ "QT": 46, "TI": 17, \\ "LC": 9, "OG": 7, \\ "PS": 5 \}\end{tabular}  & \begin{tabular}[c]{@{}c@{}}\{"LC": 5487, "OG": 511,\\ "O": 402, "PS": 58, \\ "QT": 42, "DT": 2, \\ "TI":0 \}\end{tabular} & \begin{tabular}[c]{@{}c@{}}\{"TI": 2156, "O": 47, \\ "QT": 12, "DT": 10 \}\end{tabular} \\ \hline
XPT-4M                                    & \begin{tabular}[c]{@{}c@{}}\{"PS": 13187, "O": 445, \\ "OG": 43, "LC": 9,\\ "QT": 8, "DT":2, \\ "TI": 0 \}\end{tabular}   & \begin{tabular}[c]{@{}c@{}}\{"QT": 9790, "O": 295,\\ "DT": 60, "TI": 36, \\ "LC": 8, "OG": 5,\\  "PS": 4 \}\end{tabular}   & \begin{tabular}[c]{@{}c@{}}\{"OG": 8202, "O": 593, \\ "LC": 447, "PS": 53, \\ "DT": 11, "QT": 3, \\ "TI":0 \}\end{tabular}  & \begin{tabular}[c]{@{}c@{}}\{"DT": 8079, "O": 339, \\ "QT": 38, "TI": 13, \\ "LC": 13, "OG": 4, \\ "PS": 1 \}\end{tabular} & \begin{tabular}[c]{@{}c@{}}\{"LC": 5925, "OG": 272,\\ "O": 261, "PS": 26,\\  "QT": 16, "DT": 2, \\ "TI":0 \}\end{tabular} & \begin{tabular}[c]{@{}c@{}}\{"TI": 2163, "QT": 20,\\  "O": 37,"DT": 5 \}\end{tabular}   \\ \hline
\end{tabular}
}
\caption{Model prediction results for each label in the KLUE-NER task of each model. The number below the gold label means the number of entities tagged with each label in the dev dataset. }
\label{appendixtab:ner}
\end{table*}

\begin{table*}[ht]

\centering
\scalebox{0.6}{
\renewcommand{\arraystretch}{1.2}

\begin{tabular}{c||c|c|c|c}
\hline
        Gold Label       & \textbf{per:title}                                                                                                                                         & \textbf{org:top\_members/employees}                                                                            & \textbf{per:employee\_of}                                                                                                                                 & \textbf{org:product}                                                                                                                                        \\ \hline
       The ratio of label        & 9.25\% (718)                                                                                                                                           & 6.61\% (513)                                                                                                                                            & 3.12\% (242)                                                                                                                                          & 3.03\% (235)                                                                                                                                            \\ \hline \hline
KoELECTRA-400K & \{"no\_relation": 718\}                                                                                                                           & \{"no\_relation": 513\}                                                                                                                            & \{"no\_relation": 242\}                                                                                                  & \{"no\_relation": 235\}                                                                                                                            \\ \hline
KoELECTRA-4M   & \{"no\_relation": 718\}                                                                                                                           & \{"no\_relation": 513\}                                                                                                                            & \{"no\_relation": 242\}                                                                                                                          & \{"no\_relation": 235\}                                                                                                                            \\ \hline
XPT-400K       & \begin{tabular}[c]{@{}c@{}}\{"per:titile": 634, \\ "no\_relation": 48,\\ "org:top\_members/employees": 15,\\ ...,\\ "per:product": 1\}\end{tabular} & \begin{tabular}[c]{@{}c@{}}\{"org:top\_members/employees": 306, \\ "no\_relation": 174,\\ "per:title": 6,\\ ...,\\ "per:children": 1\}\end{tabular}  & \begin{tabular}[c]{@{}c@{}}\{"per:employee\_of": 185, \\ "no\_relation": 33,\\ "per:origin": 10,\\ ...,\\ "per:alternate\_names": 1\}\end{tabular} & \begin{tabular}[c]{@{}c@{}}\{"org:product": 137, \\ "no\_relation": 75,\\ "org:members": 13,\\ ...,\\ "org:top\_members/employees": 1\}\end{tabular} \\ \hline
XPT-4M         & \begin{tabular}[c]{@{}c@{}}\{"per:titile": 648, \\ "no\_relation": 32,\\ "org:top\_members/employees": 17,\\ ...,\\ "per:spouse": 1\}\end{tabular}  & \begin{tabular}[c]{@{}c@{}}\{"org:top\_members/employees": 316, \\ "no\_relation": 145,\\ "per:title": 28,\\ ..., \\"per:children": 1\}\end{tabular} & \begin{tabular}[c]{@{}c@{}}\{"per:employee\_of": 200, \\ "no\_relation": 23,\\ "per:origin": 8,\\ ...,\\ "per:schools\_attended": 1\}\end{tabular} & \begin{tabular}[c]{@{}c@{}}\{"org:product": 156, \\ "no\_relation": 62,\\ "org:members": 10,\\ ...,\\ "org:top\_members/employees": 1\}\end{tabular} \\ \hline
\end{tabular}
}
\caption{Model prediction results for each label in the KLUE-RE task of each model. The number below the gold label means the ratio of label in the dev dataset. The number of samples tagged with each label is in parenthesis. We analyze the top-4 relations that occupy the highest ratio in the dev dataset among 29 relations (labels) except for "no\_relation".}
\label{appendixtab:re}
\end{table*}

\begin{table*}[ht]

\centering
\scalebox{0.65}{
\renewcommand{\arraystretch}{1.4}
\begin{tabular}{c|c}
\hline
\textbf{Relation}                   & \textbf{Description}                                                                                   \\ \hline \hline
per:title                  & Official or unofficial names that represent the occupational position of the specified person \\\hline
org:top\_members/employees & The representative(s) or members of the specified organization                                \\\hline
per:employee\_of           & The organization where the specified person works                                             \\\hline
per:origin                 & The origins or the nationality of the specified person                                        \\\hline
org:product                & Products or merchandise produced by the specified organization                                \\\hline
org:members                & Organizations which belong to the specified organization           \\ \hline                          
\end{tabular}
}
\caption{Part of relation description of KLUE-RE.}
\label{appendixtab:relation}
\end{table*}

\end{document}